\documentclass[runningheads]{llncs}
\usepackage{float} 
\usepackage{subcaption}
\usepackage{makecell}  
\usepackage{amsmath}
\usepackage{amsfonts}

\usepackage{rotating} 
\usepackage{booktabs} 
\usepackage{adjustbox} 
\usepackage{lscape}
\usepackage{caption} 
\usepackage{graphicx}  
\usepackage[table,xcdraw]{xcolor}  
\usepackage{longtable}
\usepackage[T1]{fontenc}
\usepackage{pgfplots}
\pgfplotsset{compat=1.18}
\usepackage{hyperref}
\usepackage[moderate]{savetrees}

\begin{document}
\title{Linking Actor Behavior to \\Process Performance Over Time}
%
%
\author{
Aurélie Leribaux\inst{1} \and
Rafael Oyamada \inst{1} \and
Johannes De Smedt\inst{1} \and
Zahra Dasht Bozorgi\inst{2} \and
Artem~Polyvyanyy\inst{2} \and 
Jochen De Weerdt\inst{1}
}
\authorrunning{A. Leribaux et al.}

\institute{
Research Centre for Information Systems Engineering, KU Leuven, Belgium
\and
The University of Melbourne, Australia
}

\maketitle              
\begin{abstract}
Understanding how actor behavior influences process outcomes is a critical aspect of process mining. Traditional approaches often use aggregate and static process data, overlooking the temporal and causal dynamics that arise from individual actors' behaviors. This limits the ability to accurately capture the complexity of real-world processes, where individual actor behavior and interactions between actors significantly shape performance. In this work, we address this gap by integrating actor behavior analysis with Granger causality to identify correlating links in time series data. We apply this approach to real-world event logs, constructing time series for actor interactions (i.e. continuation, interruption, and handovers) and process outcomes. Using Group Lasso for lag selection, we identify a small but consistently influential set of lags that capture the majority of causal influence, revealing that actors' behaviors have direct and measurable impacts on process performance, particularly throughput time. These findings demonstrate the potential of actor-centric, time series-based methods for uncovering the temporal dependencies that drive process outcomes, offering a more nuanced understanding of how individual behaviors impact overall process efficiency.
\end{abstract}

\section{Introduction}\label{sec:introduction}

Process performance analysis, which focuses on evaluating performance indicators such as waiting times, response times, throughput time (TT), bottlenecks, and case outcomes, is an essential component of process mining~\cite{vanderAalst2022}. Despite its importance, the role of actor behavior in these analyses remains relatively underexplored~\cite{10680657}. Traditional process mining generally assumes that all events in a process instance follow a single, totally ordered, and causally dependent control flow \cite{tour2021}.Therefore, they often focus on aggregate performance metrics, which collapse diverse forms of actor behavior into a single summary measure. These metrics typically overlook distinctions between, for example, uninterrupted work by the same actor, work resumed after an interruption, or handovers between different actors. This aggregation approach risks overlooking the significant influence that diverse actor behaviors can have on both performance metrics and overall process outcomes.

Recent work has taken an important step toward bridging this gap by introducing a framework for decomposing process performance based on actor behavior \cite{10680657}. The approach captures different types of actor interactions, including continuations, interruptions, and handovers, providing valuable insights into the ways actors influence performance metrics and process outcomes. However, this method primarily focuses on static performance metrics and behavioral patterns, without explicitly modeling the temporal and causal relationships between these behaviors and process outcomes or performance metrics. This limits the ability to identify time-dependent causal effects, a perspective that is valuable for understanding when certain actors' behavior influence performance metrics (e.g. delays and bottlenecks). 

To address this gap, our work extends this actor-centric perspective by incorporating time series analysis. Specifically, we apply Granger causality \cite{granger1969} to time series representing actor behaviors and process outcomes. Unlike aggregation and static case-level approaches, this method captures both the frequency and the timing of interactions, providing a more comprehensive view of the causal mechanisms at play. Our main contributions can be summarized as follows:

\begin{itemize}
    \item We present a novel framework for the integration of actor behavior classification and Granger causality analysis to uncover temporal dependencies between actor interactions and process performance.
    \item We validate our framework using real-world data from three diverse event logs (BPIC2011, BPIC2017, BPIC2019), demonstrating its applicability across multiple industries.
\end{itemize}

The remainder of this paper is structured as follows. Section \ref{sec:background} introduces the necessary terms and methods. Then, section \ref{sec:relatedwork} discusses the related work. Section \ref{sec:methodology} thoroughly describes the methodology, followed by the results in Section \ref{sec:results} and a discussion in section \ref{sec:discussion}.  Finally, Section \ref{sec:conclusion} concludes this paper.

\section{Background} \label{sec:background}

In this section, we introduce a definition of event logs and how to decompose actor behavior from them. Subsequently, we describe the construction and properties of time series for behavioral and performance contexts, including the roles of lags and a TT metric. 

\subsection{Event Logs and Actors} \label{sec:eventlogs}

A process-aware information system records execution traces in the form of event logs. An event log $E$ is a multiset of traces, where each trace represents the execution history of sequential events by a single process instance (case). In this paper, an event is a tuple denoted as
$e_i = ( c,\, a, t, r )$,
where $c$ is the identifier of the process instance,
$a \in \mathcal{A}$ is the label from the executed activity from the set of unique activities,
$t$ is the timestamp at which the event occurred,
and $r \in {R}$ is the actor (a.k.a., resource) from the set of unique actors performing the activity.
Thus, a trace \( \sigma \in E \), is a sequence of events \( \sigma_c = \langle e_1, e_2, \dots, e_n \rangle \) for a given case. 

\subsection{Actor Behavior Classification} \label{sec:behavioral-classification}

Unlike traditional approaches in process mining, where traces are regarded as isolated case sequences, our analysis emphasizes actor participation across cases. Each actor induces their own path of events, enabling the study of the interaction patterns of actors. To analyze the interaction patterns of actors, we introduce a behavioral classification based on actor transitions between consecutive events within the same case. This behavioral classification framework is inspired by the approach described in paper \cite{10680657}, generalized to a diverse set of event log data to systematically analyze actor interactions and resource usage. 
Consider two events from the same case $e_i, e_j \in \sigma_c$ where $e_i$ is followed by $e_j$ (i.e., $t_i < t_j$). Let \( r_i \) and \( r_j \) represent the actors responsible for these events, and let \( e_k \) represent an arbitrary event. We classify the actors' interaction into one of four types:

\begin{itemize}
    \item \textbf{Continuation (C):} The same actor continues the event sequence without processing any other case in between:
    \[
    f_C(e_i, e_j) = C \iff (r_i = r_j) \wedge (\not\exists e_k \in E : r_i = r_k,\, t_i < t_k < t_j)
    \]

    \item \textbf{Interruption (I):} The same actor continues the event sequence, but performs another task from a different case in between:
    \[
    f_I(e_i, e_j) = I \iff  (r_i = r_j) \wedge (\exists e_k \in E : r_i = r_k,\, t_i < t_k < t_j \mid e_k \notin \sigma_c)
    \]

    \item \textbf{Handover Idle (HI):} A different actor receives the second activity in an event sequence, and the actor was idle during the transition period:
    \[
    f_{HI}(e_i, e_j) = HI \iff (r_i \neq r_j) \wedge (\not\exists e_k \in E : r_k = r_j,\, t_i < t_k < t_j)
    \]

    \item \textbf{Handover Busy (HB):} A different actor receives the second activity in an event sequence, but the actor was already busy with other cases during the transition period:
    \[
    f_{HB}(e_i, e_j) = HB \iff (r_i \neq r_j) \wedge (\exists e_k \in E: r_k = r_j,\, t_i < t_k < t_j \mid e_k \notin \sigma_c)
    \]
\end{itemize}





This classification forms the basis for the construction of time series of behavioral indicators.

\subsection{Time Series Construction, Lags and Throughput Time} \label{sec:timeseries}

Let $T=\{1, \dots, N\}$ be a sequence of ordered time points. A \textbf{time series} is a sequence of observed values and can be defined as $\mathcal{X}=\{x_t\}_{t \in T}$. Furthermore, for each time series $\mathcal{X}=\{x_t\}_{t \in T}$ considered in this paper, we define a corresponding target time series \(\mathcal{Y} = \{y_t\}_{t \in T}\), which will be used for our Granger causality analysis.
Accordingly, a \textbf{lag} (or lagged value) shifts a time series back by a fixed number of time steps. Thus, given any integer $l \geq 1$, the lag-\emph{l} of $\mathcal{X}$ is $x_{t-l}$.
In this paper, we introduce the set of the most influential lags $L=\{1, \dots, M\}$, for $M \leq |T|$. Thus, the set of observed lagged values can be defined as $\mathcal{X}'=\{x_{t-l}\}_{l \in L}$.
To examine how actor behavior evolves over time and how it impacts performance, we will convert actor behavior and case outcome metrics into time series. 

One of the main case outcome metrics analyzed in this paper is \textbf{TT}. In process mining, TT is typically defined at the case level as the duration between the timestamp of the first event \( e_1 \) and the last event \( e_n \) in a case c: $TT(c) = t(e_n) - t(e_1)$. 
In this paper, we restrict our analysis to completed cases only, ensuring that both the first and last event timestamps are available. Ongoing cases, which are still running and have no end event recorded, are excluded from our TT calculation to maintain consistency with standard definitions and avoid biases due to incomplete traces. We first compute TT(c) at the case level. Then we group the cases by their start date (i.e., the date corresponding to $t(e_1)$) and compute the daily average throughput time as $\mathcal{TT}(d) = \frac{1}{|C_d|} \sum_{c \in C_d} TT(c)$, where $C_d$ is the set of completed cases that started on day $d$. Let $D$ denote the set of all days on which at least one case started. This results in a daily time series $\{\mathcal{TT}(d)\}_{d \in D}$, where each value reflects the average case duration for that day. This approach is straightforwardly extended to other outcome metrics tailored to each event log, which will be introduced later in this paper. Together with the behavioral time series, the $\{\mathcal{TT}(d)\}$ series allows us to test whether past actor behaviors (for example, the number of HB on day $t-5$) Granger-cause variations in TT.

This temporal modeling framework sets the stage for our methodology, which constructs time series, identifies relevant lags using regularized regression, and applies Granger causality to uncover statistically significant behavioral drivers of process outcomes. 

\section{Related Work}\label{sec:relatedwork}
Granger causality, first introduced by \cite{granger1969}, is a foundational method for detecting causal relationships in time series data. It has become a cornerstone of time series analysis, with broad applications in economics, neuroscience, and genomics. In recent years, its adaptability has extended to process mining, particularly due to advances that enable its use in complex data structures such as panel data \cite{lopez-weber-2018-testing-for-granger-causality-in-panel-data} and spatio-temporal datasets \cite{Pavasant2024}.

Within process mining, Granger causality has been applied to uncover the causal factors that drive process performance. For instance, \cite{hompes2017} extended the traditional approach to identify cause-effect relationships in business processes, focusing on key performance indicators (KPIs) such as TT and case duration per resource and per activity. While their method effectively captures the process influences, it does not delve into actor-level dynamics, which are essential for understanding how individual behaviors and interactions affect process outcomes. 

This limitation has prompted more recent research to shift toward actor-centric analyses. \cite{10680657} introduced the concept of actor behavior dimensions, emphasizing the importance of decomposing performance based on the specific actions and interactions of individual process participants. Their framework accounts for various types of interactions, such as handovers, interruptions, and continuations, that influence process efficiency. However, their approach remains relatively high-level and does not explicitly address the temporal causality of these behaviors, an aspect critical to fully understanding process dynamics. Fahland~\cite{Fahland2022} laid the groundwork for this type of analysis by introducing event knowledge graphs, which capture interactions between entities in a process. 

Complementing these perspectives,~\cite{tour2021} propose an agent-based system mining framework that models organizations as sociotechnical systems, capturing how both technical and social (human) components interact dynamically. Their approach relaxes the assumption of a single, fully ordered control flow, thereby enabling the discovery of emergent process behaviors that traditional process mining overlooks. Additionally, Bemthuis et al.\cite{Bemthuis2019} introduce an agent-based process mining architecture that integrates process mining techniques with multi-agent systems to capture complex system-wide behaviors. Similarly, the AgentSimulator framework\cite{AgentSimulator2024} enables the simulation of processes using autonomous and interacting agents, effectively capturing differences in actor behaviors, interaction preferences, and capabilities. Moreover, recent work in resource and collaboration mining emphasizes the importance of analyzing resource interactions and collaboration patterns to understand process performance~\cite{corradini2022,schubert2024,fan2017,benzin2024,jooken2023}. However, these approaches are often static rather than explicitly modeling temporal dependencies. 

Taken together, the literature shows that while Granger causality is a powerful tool for uncovering process-related dependencies, existing approaches often overlook the nuanced, temporal, and actor-specific factors that contribute to process performance. This gap motivates the need for methods that integrate temporal causality with actor behavior analysis, a direction that the proposed methodology will explore.

\section{Methodology}\label{sec:methodology}
\begin{figure}[htp]
    \centering
    \includegraphics[width=\textwidth]{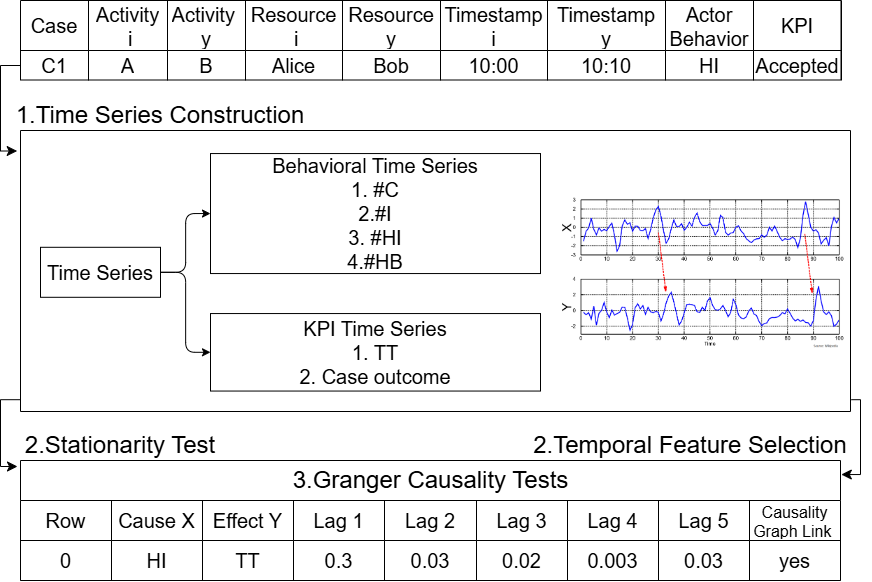}
    \caption{Methodology}
    \label{fig:workflow}
\end{figure}

Starting from preprocessed event data, our methodology comprises three main steps, as illustrated in Figure~\ref{fig:workflow}: (1) construction of time series, (2) preparation of these time series for causal analysis, and (3) Granger causality testing. The following sections describe each step in detail.

\subsection{Step 1: Time Series Construction}

The preprocessed event logs are organized into daily time series formats, where each time step corresponds to a calendar day. Events are assigned to time steps based on their recorded timestamp, and each event is treated as an atomic occurrence, meaning that activities are represented as single, instantaneous events. Two categories of time series were constructed from each event log: (1) \textbf{Behavioral time series}, which capture the aggregate frequency of each actor behavior type (C, I, HI, and HB) across the entire log as well as per user and per activity, and (2) \textbf{KPI time series}, which capture the average TT of completed cases and the fraction of cases with a predefined event log-specific outcome, both computed by grouping cases based on their start date. This design ensures that all time series share the same daily time steps and are fully aligned for time-dependent analysis and modeling.

An example can be illustrated as follows. Consider the set of actor behaviors HI. Its corresponding time series of daily observed values can be defined as $\mathcal{HI}(d)\ = \{HI_t\}_{t \in D}$.
Such definitions hence enable a clear temporal understanding of actor behaviors and KPIs, highlighting trends and anomalies across the analyzed processes. The resulting time series were visualized using the \texttt{DyLoPro} tool \cite{10.1007/978-3-031-41620-0_9}.

\subsection{Step 2: Stationarity Test and Temporal Feature Selection}

Prior to conducting Granger causality analyses, we verified that the time series data satisfied the stationarity assumption \cite{annurev}. In this context, stationarity implies that the statistical properties of the series, such as mean and variance, remain constant over time. To assess stationarity, we applied the Augmented Dickey-Fuller (ADF) test, which automatically determines the optimal lag length $|L|$ using the Schwarz Information Criterion. For any time series that failed the stationarity test, we employed first-order differencing to achieve stationarity before proceeding with the causality analysis.


Once stationarity was confirmed, the appropriate number of lagged observations was determined prior to the Granger causality test. Selecting too few lags risks omitting relevant causal dependencies, while too many can lead to overfitting and reduced interpretability \cite{annurev}. To address this, we employed Group Lasso regularization, which treats groups of coefficients collectively rather than individually. This approach hence supports the detection of consistent influences across multiple behavioral variables, allowing lags to be included or excluded as a unit across all variables. As a result, the selected set of lags can be defined as $L' \subseteq L$.

\subsection{Step 3: Granger Causality Tests}

To test whether the behavioral interactions have a causal impact on process performance, we apply Granger causality analysis~\cite{granger1969} between the behavioral and KPI time series. A time series \(\mathcal{X}\) is said to Granger-cause another series \(\mathcal{Y}\) if the past values \(\{x_{t-l}\}_{l \in L}\) help to predict \(\mathcal{Y}\) better than its past values \(\{y_{t-l}\}_{l \in L}\) alone. The test is conducted in three steps:

\begin{enumerate}
    \item \textbf{Baseline prediction (univariate model):} A linear autoregressive model is fitted to \(\mathcal{Y}\), using only its own past $L$ values:
    \[
    \mathcal{Y}_{t} = \sum_{l=1}^L \alpha_l y_{t-l} + \epsilon_t \,.
    \]
    
    \item \textbf{Augmented prediction (bivariate model):} A second model includes both the past values of \(\mathcal{Y}\) and \(\mathcal{X}\):
    \[
    \mathcal{Y}_{t} = \sum_{l=1}^L \alpha_l y_{t-l} + \sum_{l=1}^L \beta_l x_{t-l} + \eta_t \,.
    \]

    \item \textbf{Statistical comparison:} The prediction errors (\(\epsilon_t\) and \(\eta_t\)) of the two models are compared using an F-test based on the residual sum of squares (SSR). This test evaluates whether including the lagged values of \(\mathcal{X}\) significantly improves the prediction of \(\mathcal{Y}\). The F-statistic is computed as:
\[
F = \frac{(\text{SSR}_{\text{univariate}} - \text{SSR}_{\text{bivariate}})/L}{\text{SSR}_{\text{bivariate}} / (T - 2L - 1)} \,,
\]
where \(L\) is the number of lags, and \(T\) is the number of observations. A low p-value from the F-test indicates that \(\mathcal{X}\) Granger-causes \(\mathcal{Y}\).
\end{enumerate}


We perform these tests over the selected lags (Step~2) to determine whether past behavior frequencies Granger cause changes in performance metrics. For each pair of time series, a relationship is considered significant if the p-value falls below 0.05 at any of the tested lags. A causal graph is then constructed by including only those pairs with at least one significant lag, where a directed edge denotes that \( \mathcal{X} \) Granger causes \( \mathcal{Y} \) at one or more lagged observations.

\section{Evaluation}\label{sec:results}
We evaluate our approach using three real-life event logs from the Business Process Intelligence Challenge (BPIC) series. These datasets span multiple domains and contain detailed information about both process execution and actor involvement, making them well-suited for analyzing the causal impact of actor behavior on performance outcomes. We implemented the approach in Python, and the code is available on our GitHub repository\footnote{\url{https://github.com/aurelieleribaux-1/ActorBehaviorGranger}}.The following sections explain the setup and the results for each dataset.

\subsection{Setup}\label{sec:setup}

The first dataset, BPIC 2011 (Hospital log)\footnote{\url{https://data.4tu.nl/articles/dataset/Real-life_event_logs_-_Hospital_log/12716513}}, captures patient treatment processes in a Dutch hospital. The second dataset, BPIC 2017\footnote{\url{https://data.4tu.nl/articles/dataset/BPI_Challenge_2017/12696884}}, documents the handling of loan applications at a Dutch financial institution and serves as a more structured and complete version of the earlier BPIC 2012 dataset. The third dataset, BPIC 2019 \footnote{\url{https://data.4tu.nl/datasets/46a7e15b-10c7-4ab2-988d-ee67d8ea515a}}, describes a purchase-to-pay process in a coatings and paint manufacturing company.

For each log, we first identified actor behavior by analyzing every pair of consecutive events within a case. To support outcome-oriented analysis, we then defined KPIs tailored to each dataset. For BPIC 2011, a binary outcome variable was engineered to indicate whether a surgical procedure occurred in a case. This was determined by scanning activity labels for medically relevant Dutch keywords such as \texttt{'operatie'}, \texttt{'resectie'}, \texttt{'excisie'}, and \texttt{'extirpaties'}. For BPIC 2017, the process outcome was derived from the final event in each case, which indicates whether a loan application was accepted, rejected, or cancelled, based on event labels prefixed with \texttt{O\_} (e.g., \texttt{O\_ACCEPTED}). The BPIC 2019 dataset already contains a boolean attribute indicating whether the ordered goods were received, so no additional outcome engineering was necessary.

We constructed two categories of time series (i.e. behavioral and KPI). For all three event logs, we included the frequency of the behaviors and the time series of TT. Additionally, we included the frequency of the behavior types per user and per activity and the fraction of accepted loans (\%Acc) in BPIC 2017, the fraction of goods received (\%GR) in BPIC 2019, and the fraction of cases involving surgical procedures (\%OP)in BPIC 2011. To identify the most influential lags $L'$, we tested a broad range of regularization parameters over a maximum lag window of 22 days. Specifically, we varied the group regularization parameter ($\lambda_g$), which controls the number of active lag groups, across the set $\lambda_g=\{0.01, 0.1, 0.5, 1, 5, 10\}$. In parallel, the L1 regularization parameter ($\lambda_1$), which promotes sparsity within those selected groups, was varied over $\lambda_1=\{0.0, 0.01, 0.1, 0.5, 1\}$. From the resulting models, we selected the 5 most frequently chosen lags to serve as candidates in the subsequent Granger causality tests.

The following hypotheses were constructed to guide the evaluation and results:
\begin{enumerate}
    \item \textbf{Hypothesis 1}: Historical behavioral trends $\mathcal{X}$ at lag $\{x_{t{-}l}\}_{l \in L'}$ significantly cause changes in process performance outcomes $\mathcal{Y}$ at day $t$.
    \item \textbf{Hypothesis 2}: If $\{x_{t{-}l}\}_{l \in L'} \rightarrow y_t$ exhibits significant causal influence, then $\{y_{t{-}l}\}_{l \in L'} \rightarrow x_t$ should not show significant causality, to ensure an one-directional effect.
\end{enumerate} 

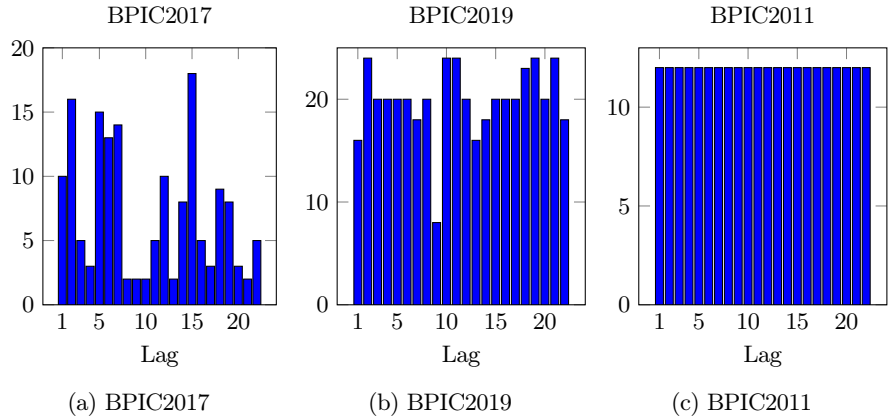
\begin{figure}[htbp]
    \centering

    \begin{subfigure}[b]{0.32\linewidth}
        \centering
        \begin{tikzpicture}
        \begin{axis}[
            width=1.2\linewidth,
            height=5cm,
            title={BPIC2017},
            xlabel={Lag},
            ymin=0, ymax=20,
            xtick={1,5,10,15,20},
            bar width=3pt,
        ]
        \addplot[ybar, fill=blue] coordinates {
            (1,10) (2,16) (3,5) (4,3) (5,15) (6,13) (7,14) (8,2) (9,2) (10,2)
            (11,5) (12,10) (13,2) (14,8) (15,18) (16,5) (17,3) (18,9)
            (19,8) (20,3) (21,2) (22,5)
        };
        \end{axis}
        \end{tikzpicture}
        \caption{BPIC2017}
    \end{subfigure}
    %
    \begin{subfigure}[b]{0.32\linewidth}
        \centering
        \begin{tikzpicture}
        \begin{axis}[
            width=1.25\linewidth,
            height=5cm,
            title={BPIC2019},
            xlabel={Lag},
            ymin=0, ymax=25,
            xtick={1,5,10,15,20},
            bar width=3pt,
        ]
        \addplot[ybar, fill=blue] coordinates {
            (1,16) (2,24) (3,20) (4,20) (5,20) (6,20) (7,18) (8,20) (9,8) (10,24)
            (11,24) (12,20) (13,16) (14,18) (15,20) (16,20) (17,20) (18,23)
            (19,24) (20,20) (21,24) (22,18)
        };
        \end{axis}
        \end{tikzpicture}
        \caption{BPIC2019}
    \end{subfigure}
    %
    \begin{subfigure}[b]{0.32\linewidth}
        \centering
        \begin{tikzpicture}
        \begin{axis}[
            width=1.25\linewidth,
            height=5cm,
            title={BPIC2011},
            xlabel={Lag},
            ymin=0, ymax=13,
            xtick={1,5,10,15,20},
            bar width=3pt,
        ]
        \addplot[ybar, fill=blue] coordinates {
            (1,12) (2,12) (3,12) (4,12) (5,12) (6,12) (7,12) (8,12) (9,12)
            (10,12) (11,12) (12,12) (13,12) (14,12) (15,12) (16,12)
            (17,12) (18,12) (19,12) (20,12) (21,12) (22,12)
        };
        \end{axis}
        \end{tikzpicture}
        \caption{BPIC2011}
    \end{subfigure}

    \caption{l (1-22) selection frequency per dataset across all $(\lambda_g, \lambda_1)$ combinations.}
    \label{fig:all_bpic_lags}
\end{figure}

\subsection{Results}
In this section, we present the results related to the proposed hypotheses across all datasets. We begin by introducing the lag selection procedure. \autoref{fig:all_bpic_lags} shows the frequency with which each lag was selected by the Group Lasso method. For the BPIC2017 and BPIC2019 datasets, we selected the five most frequently chosen lags: $\{2, 5, 6, 7, 15\}$ for BPIC2017 and $\{2, 10, 11, 19, 2\}$ for BPIC2019. For BPIC2011, where all lags were selected with similar frequency, we used the first five lags: $\{1, 2, 3, 4, 5\}$.

\subsubsection{BPIC2017}
The results of the Granger causality analysis are presented in Table~\ref{tab:results-bpic2017}, revealing noteworthy insights.
Consistent with Hypothesis 1, we observe strong and recurring causal influence of \textbf{C}, \textbf{HB}, and \textbf{HI}, on \textbf{TT}. These effects are significant across all $l$. Moreover, the influence from \textbf{I} becomes significant l=15, suggesting a delayed impact on TT. These results support the hypothesis that behavioral trends are associated with subsequent changes in TT. While the influence on \textbf{\%Acc} is generally weaker and sporadic, there are still notable findings. In particular, the significant relationship from \textbf{HB,HI} to \textbf{\%Acc} at l=7,15, indicates that H may have a delayed influence on loan acceptance.

\begin{figure}[htb]
    \centering
    \begin{subfigure}[t]{0.49\textwidth}
        \centering
        \includegraphics[width=\textwidth]{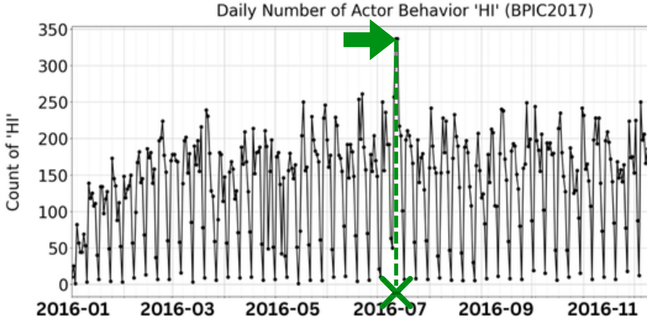}
        \caption{Daily Number of HI Transitions}
        \label{fig:numberHI}
    \end{subfigure}%
    \hfill
    \begin{subfigure}[t]{0.49\textwidth}
        \centering
        \includegraphics[width=\textwidth]{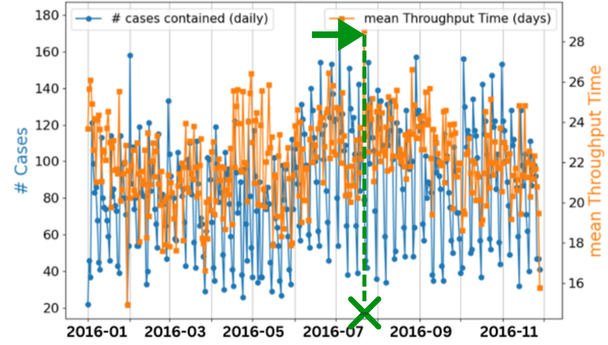}
        \caption{Mean TT cases}
        \label{fig:meanTT}
    \end{subfigure}
    
    \caption{Comparison of daily actor behavior HI and mean TT for BPIC2017.}
    \label{fig:bpic2017_combined}
\end{figure}

\begin{figure}[htp]
    \centering
     \includegraphics[width=\textwidth]{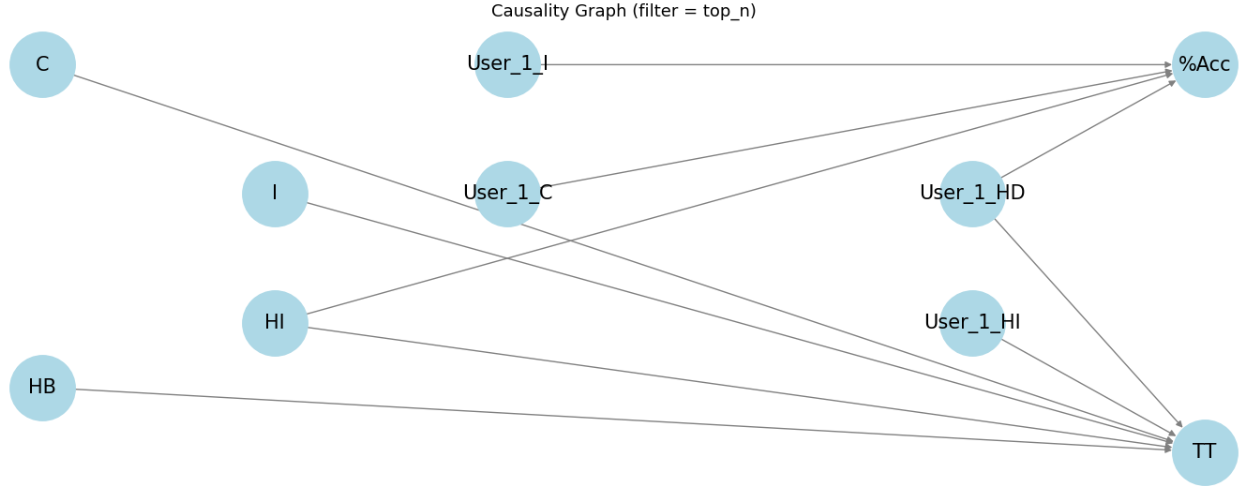}    
     \caption{Causality graph}
    \label{fig:graph}
\end{figure}

Figure \ref{fig:graph} presents a causality graph of the top 10 most significant influences for BPIC2017, highlighting how actor behavior affects process outcomes. For instance, User 1’s handovers to other resources significantly impact both TT and \%Acc. Due to space limitations, similar detailed analyses of the remaining datasets are omitted. The time series in Figure \ref{fig:bpic2017_combined} visually supports the Granger results from the graph and Table \ref{tab:results-bpic2017}, with peaks in HI often preceding rises in TT. For example, a HI peak in July 2016 (left) is followed by a TT peak (right), illustrating their lagged relationship.

Hypothesis 2, which posits an asymmetry in causal relationships, was satisfied in 55.55\% of the significant cases. This support suggests that, in most cases, causal links demonstrate clear directionality.


\begin{table}[htb]
\centering
\scriptsize
\caption{BPIC2017 Granger causality p-values (selected l). Each cell shows $X \rightarrow Y$ with $Y \rightarrow X$ in parentheses \textbf{only if} $X \rightarrow Y$ is significant. Underlined = significant ($p < 0.05$).}
\resizebox{\textwidth}{!}{%
\begin{tabular}{lcccccccc}
\toprule
\textbf{l} &
\textbf{C $\rightarrow$ \%Acc.} &
\textbf{C $\rightarrow$ TT} &
\textbf{HB $\rightarrow$ \%Acc.} &
\textbf{HB $\rightarrow$ TT} &
\textbf{HI $\rightarrow$ \%Acc.} &
\textbf{HI $\rightarrow$ TT} &
\textbf{I $\rightarrow$ \%Acc.} &
\textbf{I $\rightarrow$ TT} \\
\midrule
2 & 
0.858 & 
\underline{0.001} (0.089) & 
0.855 & 
\underline{0.002} (0.050) & 
0.619 & 
\underline{0.001} (0.061) & 
0.991 & 
0.180 \\
\midrule
5 & 
0.799 & 
\underline{0.002} (\underline{0.009}) & 
0.652 & 
\underline{0.004} (\underline{0.009}) & 
0.953 & 
\underline{0.008} (\underline{0.036}) & 
0.998 & 
0.101 \\
\midrule
6 & 
0.074 & 
\underline{0.005} (\underline{0.015}) & 
0.076 & 
\underline{0.010} (0.613) & 
0.166 & 
\underline{0.022} (0.950) & 
0.556 & 
0.259 \\
\midrule
7 & 
0.075 & 
\underline{0.006} (\underline{0.014}) & 
\underline{0.044} (\underline{0.004}) & 
\underline{0.012} (0.684) & 
0.210 & 
\underline{0.026} (0.973) & 
0.652 & 
0.153 \\
\midrule
15 & 
0.254 & 
\underline{0.005} (\underline{0.003}) & 
0.145 & 
\underline{0.007} (0.133) & 
\underline{0.040} (0.050) & 
\underline{0.038} (0.559) & 
0.299 & 
\underline{0.024} (\underline{0.027}) \\
\bottomrule
\end{tabular}
}
\label{tab:results-bpic2017}
\end{table}
\vspace{-1cm}

\subsubsection{BPIC2019}



Granger causality results (Table~\ref{tab:results-bpic2019}) support hypothesis 1: \textbf{HB} $\rightarrow$ \textbf{TT} is significant at lag 21 ($p = 0.039$), suggesting a delayed effect of busy handovers, while \textbf{HI} $\rightarrow$ \textbf{TT} is significant at lag 2 ($p = 0.031$), indicating a more immediate impact during idle periods. Additionally, \textbf{I} $\rightarrow$ \textbf{\%OP} shows significance at lags 19 ($p = 0.025$) and 21 ($p = 0.002$), suggesting that interruptions may disrupt process accuracy or completion.

With respect to Hypothesis 2, we find that the asymmetric causality condition is satisfied in 50\% of the significant cases, indicating a moderate level of support for the directionality assumption.

\begin{table}[htb]
\centering
\scriptsize
\caption{BPIC2019 Granger causality p-values (selected l). Each cell shows $X \rightarrow Y$ with $Y \rightarrow X$ in parentheses \textbf{only if} $X \rightarrow Y$ is significant. Underlined = significant ($p < 0.05$).}
\resizebox{\textwidth}{!}{%
\begin{tabular}{lcccccccc}
\toprule
\textbf{l} &
\textbf{C $\rightarrow$ \%GR} &
\textbf{C $\rightarrow$ TT} &
\textbf{HB $\rightarrow$ \%GR} &
\textbf{HB $\rightarrow$ TT} &
\textbf{HI $\rightarrow$ \%GR} &
\textbf{HI $\rightarrow$ TT} &
\textbf{I $\rightarrow$ \%GR} &
\textbf{I $\rightarrow$ TT} \\
\midrule
2  & 
0.091 & 
0.063 & 
0.081 & 
0.079 & 
0.170 & 
\underline{0.031} (0.520) & 
0.177 & 
0.368 \\
\midrule
10 & 
0.167 & 
0.063 & 
0.199 & 
0.318 & 
0.159 & 
0.087 & 
0.626 & 
0.252 \\
\midrule
11 & 
0.204 & 
0.059 & 
0.194 & 
0.229 & 
0.098 & 
0.080 & 
0.302 & 
0.205 \\
\midrule
19 & 
\underline{0.031} (\underline{6e-4}) & 
0.163 & 
0.080 & 
0.083 & 
0.433 & 
0.121 & 
\underline{0.025} (0.213) & 
0.575 \\
\midrule
21 & 
0.061 & 
0.064 & 
\underline{0.021} (\underline{3e-7}) & 
\underline{0.039} (\underline{5e-3}) & 
0.569 & 
0.121 & 
\underline{0.002} (0.093) & 
0.526 \\
\bottomrule
\end{tabular}
}
\label{tab:results-bpic2019}
\end{table}

 \subsubsection{BPIC2011 Hospital Log}

All $l$ may have causal influence, with detailed Granger causality results presented in Table~\ref{tab:results-bpic2011}. Analyzing these $l$ reveals strong support for Hypothesis 1, with nearly all behavior-to-performance pathways demonstrating significant Granger causality. \textbf{C}, \textbf{I}, and both types of \textbf{H} significantly influence \textbf{TT} and the \textbf{\%OP} at higher $l$. These findings indicate that case-level actor dynamics have a clear and immediate effect on clinical process performance. For example, if actors frequently switch tasks or handle multiple cases, this can impact their efficiency, directly affecting the TT and/or the fraction of operations completed. 

Regarding Hypothesis 2, the assumption of causal asymmetry is satisfied in 53.33\% of the significant cases. This indicates that, in more than half of the observed relationships, the causality appears to be directional rather than bidirectional. 


\begin{table}[htb]
\centering
\scriptsize
\caption{BPIC2011 Granger causality p-values (selected l). Each cell shows $X \rightarrow Y$ with $Y \rightarrow X$ in parentheses \textbf{only if} $X \rightarrow Y$ is significant. Underlined = significant ($p < 0.05$).}
\resizebox{\textwidth}{!}{%
\begin{tabular}{lcccccccc}
\toprule
\textbf{l} &
\textbf{C $\rightarrow$ \%OP} &
\textbf{C $\rightarrow$ TT} &
\textbf{HB $\rightarrow$ \%OP.} &
\textbf{HB $\rightarrow$ TT} &
\textbf{HI $\rightarrow$ \%OP} &
\textbf{HI $\rightarrow$ TT} &
\textbf{I $\rightarrow$ \%OP} &
\textbf{I $\rightarrow$ TT} \\
\midrule
1 & 
0.465 & 
\underline{0.006} (0.901) & 
0.281 & 
\underline{1.7e-5} (0.572) & 
0.652 & 
\underline{0.001} (0.258) & 
0.783 & 
\underline{2e-4} (0.911) \\
\midrule
2 & 
0.403 & 
0.0889 & 
0.515 & 
\underline{0.001} (0.258) & 
0.901 & 
\underline{0.017} (0.869) & 
0.275 & 
\underline{4e-5} (0.078) \\
\midrule
3 & 
\underline{0.015} (\underline{2e-5}) & 
\underline{0.004} (0.360) & 
\underline{1e-3} (0.050) & 
\underline{1e-5} (0.904) & 
0.076 & 
\underline{0.003} (0.670) & 
\underline{1e-6} (0.295) & 
\underline{3.1e-14} (\underline{0.002}) \\
\midrule
4 & 
\underline{0.030} (\underline{1e-5}) & 
\underline{9.3e-5} (\underline{0.026}) & 
\underline{1.4e-5} (\underline{0.009}) & 
\underline{1.1e-8} (0.838) & 
\underline{0.042} (\underline{1e-4}) & 
\underline{5e-5} (0.125) & 
\underline{1e-6} (\underline{2.5e-4}) & 
\underline{6.4e-13} (0.128) \\
\midrule
5 & 
\underline{2.4e-4} (\underline{1.1e-5}) & 
\underline{1e-6} (\underline{9e-4}) & 
\underline{1.3e-5} (\underline{0.014}) & 
\underline{3.5e-8} (0.887) & 
\underline{2.8e-4} (\underline{1.1e-5}) & 
\underline{1.2e-7} (\underline{0.002}) & 
\underline{2e-6} (\underline{0.001}) & 
\underline{3.4e-10} (\underline{0.002}) \\
\bottomrule
\end{tabular}
}
\label{tab:results-bpic2011}
\end{table}

\section{Discussion} \label{sec:discussion}
The results presented in Section~\ref{sec:results} reveal several important insights in the causal influence of different actor behavior dynamics towards process outcomes across different event logs. The most consistent finding is the strong causal influence of handovers, both HB and HI, on TT. This is evident across multiple datasets and lag configurations, intuitively implying that a higher number of handovers increases the TT, and vice versa. Moreover, several KPIs' outcomes show promising potential, like in BPIC2017, where handovers lead to a long-term dependency on \%Acc. The one-directionality was satisfied in all datasets. These findings underscore the role of actor behavior in process efficiency, especially in complex domains like healthcare, finance, and procurement. Moreover, the approach scales to large event logs and can be applied in real-world settings using standard event log data, enabling resource-level behavior to inform system-level (e.g., TT) and case-level (e.g., remaining time) predictions.

Despite these promising results, several limitations should be noted. First, we rely on Granger causality, which does not account for latent confounding effects and cannot capture multivariate or non-linear relationships. Additionally, as discussed in Section~\ref{sec:relatedwork}, no other time-dependent frameworks focusing specifically on actor behavior exist for direct comparison, limiting our ability to benchmark our framework. Finally, the classification of actor behaviors itself is dependent on the precision of the event log timestamps and task definitions. This can lead to misclassifications if timestamps are coarse, tasks are imprecisely defined, or actors handle multiple tasks simultaneously. The current framework assumes that actors only work on a single task at a time and treats tasks spanning multiple cases as separate, potentially overestimating busy handovers. 

Future work should focus on integrating more flexible, non-linear causal inference methods and extending the framework to handle multi-tasking actors and continuous tasks more effectively. 

\section{Conclusion}  \label{sec:conclusion}
In this paper, we examined how actor behavior influences process performance across loan application  (BPIC 2017), purchase-to-pay (BPIC 2019), and hospital processes (BPIC 2011) using Granger causality. 
Actor behavior is decomposed into continuation, interruption, handover idle, and handover busy. Time series are constructed on the number of behaviors and  KPIs of the process, between which Granger causalities are tested.
Collectively, the findings underscore the critical role of actor behavior in process efficiency. Handovers mainly have distinct, lag-dependent impacts on TT and other process outcomes, reflecting complex interactions between task transfers, actor availability, and coordination. This understanding can improve predictive modeling by anticipating bottlenecks, optimizing resource allocation, and improving overall process performance. Future work could explore how these (handover) behaviors help improve predictive modeling or how they interact with other factors, like case complexity, resource availability, and workload distribution, to further refine process insights.

\bibliographystyle{splncs04}
\bibliography{bibliography}   
\end{document}